\documentclass{article}
\usepackage[table]{xcolor}
\usepackage{microtype}
\usepackage{graphicx}
\usepackage{subfigure}
\usepackage{booktabs} 
\usepackage{comment}
\usepackage{hyperref}

\usepackage[accepted]{conf}

\usepackage{amsmath}
\usepackage{amssymb}
\usepackage{mathtools}
\usepackage{amsthm}

\usepackage[capitalize,noabbrev]{cleveref}

\theoremstyle{plain}

\theoremstyle{definition}

\theoremstyle{remark}

\usepackage[disable,textsize=tiny]{todonotes}

\usepackage{float}
\usepackage{bm}
\usepackage{xspace}
\usepackage[T1]{fontenc}
\usepackage{multirow}
\usepackage{makecell}
\usepackage{array}

\newcommand{\onedot}{.\@\xspace}
\newcommand{\eg}{\emph{e.g}\onedot}

\newcommand{\ie}{\emph{i.e}\onedot}

\newcommand{\etc}{\emph{etc}\onedot}

\newcommand{\etal}{\emph{et al}\onedot}

\definecolor{myblue}{RGB}{66,133,244}
\definecolor{mygreen}{RGB}{51,168,83}
\definecolor{myyellow}{RGB}{251,188,3}
\definecolor{myred}{RGB}{234,67,53}
\definecolor{mygrey}{RGB}{95,99,104}

\begin{document}

\twocolumn[
\icmltitle{Trust-Aware Diversion for Data-Effective Distillation}

\icmlsetsymbol{equal}{*}

\begin{icmlauthorlist}
\icmlauthor{Zhuojie Wu}{UQ}
\icmlauthor{Yanbin Liu}{AUT}
\icmlauthor{Xin Shen}{UQ}
\icmlauthor{Xiaofeng Cao}{JLU}
\icmlauthor{Xin Yu}{UQ}

\end{icmlauthorlist}

\centerline{
\textsuperscript{1}The University of Queensland \quad
\textsuperscript{2}Auckland University of Technology \quad
\textsuperscript{3}Jilin University
}

\icmlaffiliation{UQ}{The University of Queensland}
\icmlaffiliation{AUT}{Auckland University of Technology}
\icmlaffiliation{JLU}{Jilin University}

\vskip 0.3in
]

\begin{abstract}

Dataset distillation compresses a large dataset into a small synthetic subset that retains essential information. 
Existing methods assume that all samples are perfectly labeled, limiting their real-world applications where incorrect labels are ubiquitous. 
These mislabeled samples introduce untrustworthy information into the dataset, which misleads model optimization in dataset distillation.
To tackle this issue, we propose a \textbf{Trust-Aware Diversion (TAD)} dataset distillation method.
Our proposed TAD introduces an iterative dual-loop optimization framework for data-effective distillation.
Specifically, the outer loop divides data into trusted and untrusted spaces, redirecting distillation toward trusted samples to guarantee trust in the distillation process.
This step minimizes the impact of mislabeled samples on dataset distillation.
The inner loop maximizes the distillation objective by recalibrating untrusted samples, thus transforming them into valuable ones for distillation.
This dual-loop iteratively refines and compensates for each other, gradually expanding the trusted space and shrinking the untrusted space.
Experiments demonstrate that our method can significantly improve the performance of existing dataset distillation methods on three widely used benchmarks (CIFAR10, CIFAR100, and Tiny ImageNet) in three challenging mislabeled settings (symmetric, asymmetric, and real-world).

\end{abstract}

\section{Introduction}

\begin{figure}[t]
    \centering
    \includegraphics[width=0.95 \linewidth]{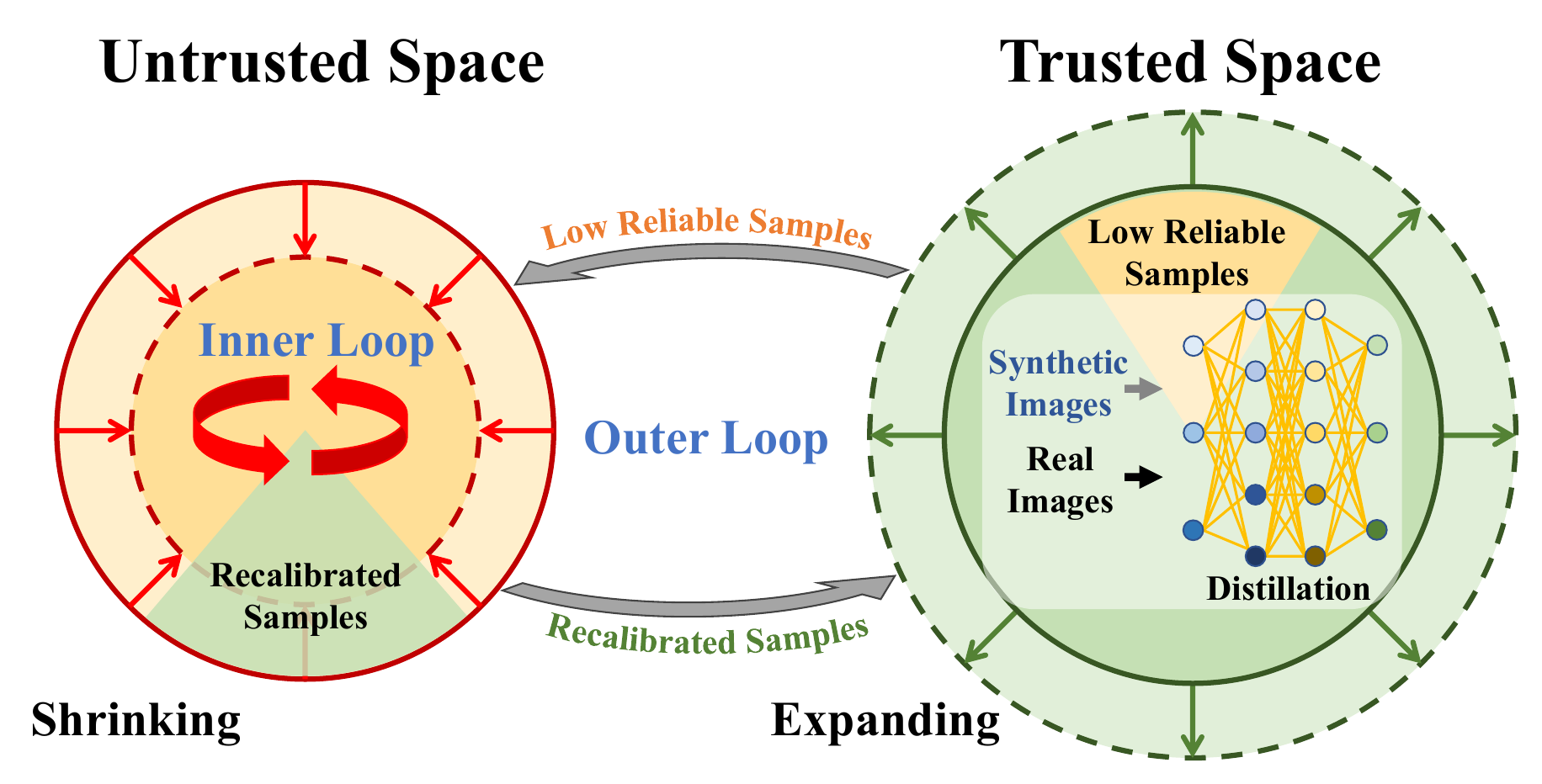}
    \vspace{-1em}
    \caption{
    Illustration of our proposed Trust-Aware Diversion (TAD) dataset distillation method.
    The outer loop separates data into trusted and untrusted spaces, rerouting distillation toward trusted samples.
    The inner loop recalibrates untrusted samples and transposes them to the trusted space.
    Through iterative interactions, the trusted space expands while the untrusted space shrinks, improving dataset distillation under noise conditions.
    }
    \vspace{-1em}
    \label{fig1}
\end{figure}

In an era of massive datasets, data-efficient learning is pivotal for achieving high performance with a limited computation budget. 
Dataset Distillation (DD) presents a promising solution by synthesizing a small amount of highly informative data that summarizes large volumes of real data, allowing models to maintain high performance with less data~\cite{zhao2021datasetcondensation, cazenavette2022distillation, deng2024exploiting}.

The current success of DD relies on the assumption that all labels are completely correct. 
However, erroneously labeled data is ubiquitous in real-world scenarios. 
State-of-the-art DD methods~\cite{du2023minimizing, guo2024lossless, liu2024dataset} usually adopt a trajectory matching strategy, which aligns the parameters trained on distilled synthetic data (\ie, student model) with the parameters trained on real data (\ie, expert model). 
With the existence of noisy labels, an expert model tends to produce a misrepresented trajectory caused by the incorrectly labeled data. 
Consequently, if a student model attempts to match this expert trajectory, it would produce inferior synthetic data and suffer overall performance degradation.

In this paper, we explore a more realistic scenario, \ie, Dataset Distillation with Noisy Labels (DDNL).
Notably, noisy labels introduce a unique challenge to DD: how to maintain a trustworthy distillation process without being misled by noisy samples.
We observed that existing methods struggle to identify mislabeled or noisy data perfectly~\cite{li2022neighborhood, zhao2022centrality, karim2022unicon}.
Thus, when these methods are applied, noisy data inevitably propagates into the dataset distillation, amplifying errors and compromising the performance of the distillation.

\begin{figure}[t]
    \centering
    \includegraphics[width=.99 \linewidth]{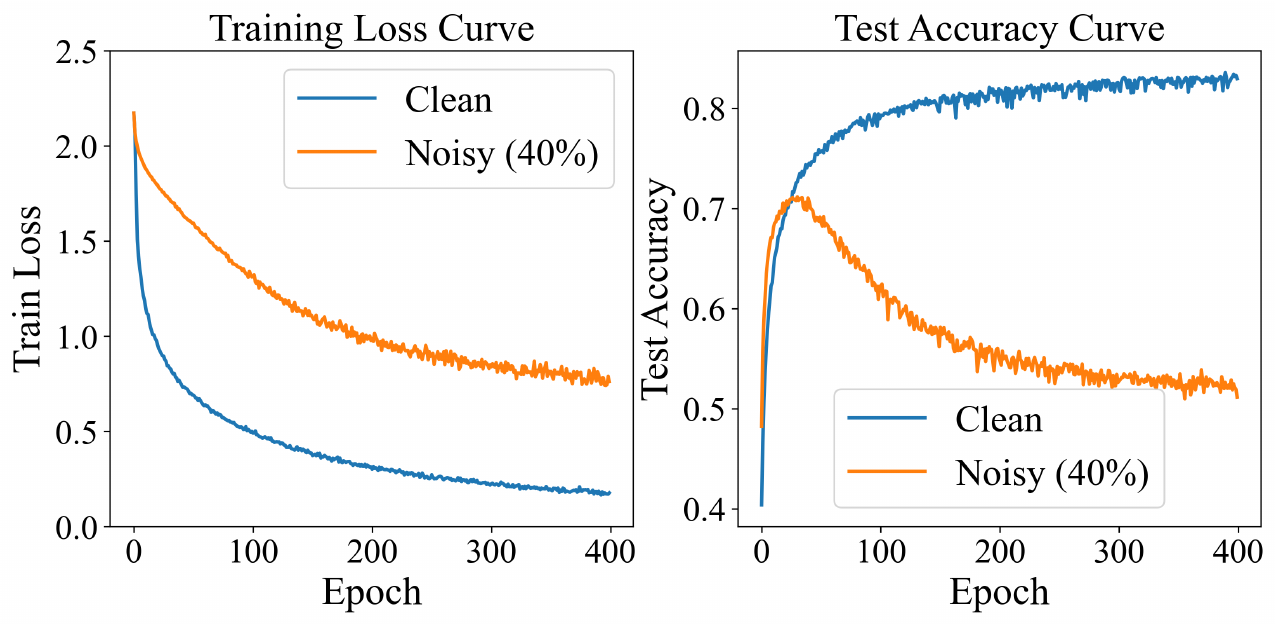}
    \vspace{-1em}
    \caption{
    Training loss and test accuracy curves of an expert model on CIFAR-10 with clean and 40\% symmetric noisy labels. 
    Noise impedes convergence and significantly degrades performance.
   }
   \vspace{-1em}
    \label{fig2}
\end{figure}

To tackle this issue, we propose a \textbf{Trust-Aware Diversion (TAD)} dataset distillation method, which introduces an iterative dual-loop optimization framework (\ie, an outer loop and an inner loop) for data-effective distillation. 
The outer loop is in charge of diverting data into trusted and untrusted spaces to ensure trustworthy distillation, while the inner loop recalibrates untrusted samples, refining them into valuable data for distillation.
Through this iterative interaction, TAD can reliably refine the distinction between trusted and untrusted samples, progressively mitigating the impact of mislabeled data on data distillation.

To be specific, the outer loop separates data into trusted and untrusted spaces, and redirects distillation towards trusted samples, thus ensuring a trustworthy distillation process.
The memorization effect~\cite{arpit2017closer,yao2020searching} of deep neural networks (DNNs) shows that DNNs first learn clean samples and then noisy samples. 
In addition, we observed a clear gap in loss values between clean and noisy samples during training. 
Clean samples usually have smaller losses, while noisy samples typically exhibit higher losses, as shown in Fig.~\ref{fig2}.
Motivated by this, data can be interpreted as a mixture of two distinct sample categories (\ie, trusted and untrusted).
Thus, we employ the class-wise mean posterior probability for the division criterion.
In addition, we introduce a consistent regularization term that avoids overfitting to the wrongly divided samples, ensuring trusted samples contribute to distillation effectively.

Although data has been partitioned into trusted and untrusted spaces in the outer loop, some noisy samples may still be mistakenly classified as trusted ones.
To address this issue, we introduce a reliability score in the inner loop to quantify the trustworthiness of each sample in the trusted space, providing a fine-grained evaluation of reliability.
Since the distilled synthetic data captures a representative pattern of real data~\cite{li2024prioritize, guo2024lossless}, we measure the reliability of samples in the trusted space by computing their Mahalanobis distance~\cite{de2000mahalanobis} to the class distributions.
Based on the reliability score, we rank trusted samples and select the most reliable ones from each class.
Then, these highly-reliabile samples can be used to sieve less reliable trusted samples by measuring their similarities. 
Moreover, they can be employed to recall samples from the untrusted space, thus fully exploiting all the samples.
The inner loop compensates for the outer loop, thus further preventing the propagation of unreliable information into dataset distillation.
Therefore, this iterative interaction between the inner and outer loops continuously refines the data partitioning, progressively expanding the trusted space while shrinking the untrusted space.

To the best of our knowledge, we are the first to investigate the Dataset Distillation with Noisy Labels (DDNL). 
Extensive experiments validate the superior performance of our proposed \textbf{Trust-Aware Diversion (TAD)} dataset distillation method.
For instance, on CIFAR-100 with 40\% symmetric mislabeled data, our proposed TAD consistently outperforms ATT~\cite{liu2024dataset}.
When Images Per Class (IPC) is 10, TAD achieves 41.5\% accuracy, significantly outperforming ATT (32.6\%) and the two-stage baseline (36.5\%), which first denoises the data followed by applying dataset distillation.
Similarly, when IPC is 50, TAD attains 44.2\% accuracy, outperforming ATT (37.1\%) and the two-stage baseline (41.0\%).

\section{Related Work}

\noindent{\textbf{Dataset Distillation.}} 
Dataset distillation aims to create a compact synthetic dataset that preserves the essential information in the large-scale original dataset, making it more efficient for training while achieving performance comparable to the original dataset.
Dataset distillation was initially introduced by Wang~\etal ~\cite{wang2018dataset}, drawing inspiration from Knowledge Distillation~\cite{hinton2015distilling}.
Subsequent work has explored various methods to enhance the effectiveness of dataset distillation, focusing on different strategies to match the information in the original and synthetic datasets. 
\textit{Gradient matching} aims to align gradients by performing a one-step distance-matching process between the network trained on the original dataset and the same network trained on the synthetic data~\cite{zhao2021datasetcondensation, lee2022dataset, zhao2021dataset}.
\textit{Distribution matching} directly aligns the data distributions of original and synthetic datasets through a single-level optimization process, offering an efficient yet effective strategy for dataset distillation~\cite{sajedi2023datadam, liu2023wasserstein, zhao2023improved, deng2024exploiting}.
\textit{Training trajectory matching} aligns the training trajectories of models trained on the original versus synthetic data, allowing multi-step optimization to capture nuanced information about how model parameters evolve during training~\cite{cazenavette2022distillation, yang2024nsd, du2023minimizing, guo2024lossless, liu2024dataset}. 
Other concurrent works have improved dataset distillation baselines with various approaches, including soft labels~\cite{xiao2024soft, sucholutsky2021soft, qin2024label}, decoupled distillation~\cite{yin2023sre2l, shao2024gvbsm, sun2024rded}, data parameterization~\cite{wei2023sparse, son2024fyi, kim2022dataset},~\etc.
Dataset distillation has enabled various applications including neural architecture search~\cite{medvedev2021tabular, cui2022dc, zhao2021dataset}, continual learning~\cite{sangermano2022sample, yang2023efficient, gu2024ssd}, privacy protection~\cite{chung2024backdoor, li2023sharing, li2024infodist},~\etc.

\noindent{\textbf{Practical Dataset Distillation.}}  
With the rapid advancement of deep learning techniques and optimization methods, dataset distillation has achieved remarkable progress.
However, these methods assume that all labels are completely correct.
This assumption may not hold in real-world scenarios, where mislabeled or noisy data is ubiquitous.

Imperfect data can significantly impact the effectiveness of dataset distillation.
The synthetic dataset generated from such data often inherits the inconsistencies and errors of the original dataset. 
As a result, it struggles to accurately capture essential information from the original dataset.
Specifically, surrogate-based methods~\cite{zhao2021datasetcondensation, zhao2021dataset, du2023minimizing, guo2024lossless, liu2024dataset} rely on accurate gradients derived from the original dataset.
Noisy or mislabeled data can distort these gradients, leading to incorrect optimization.
In addition, distribution-based methods~\cite{sajedi2023datadam, zhao2023improved, yin2023sre2l, sun2024rded} aim to match distributions between the original and synthetic datasets in a randomly initialized or pretrained network.
However, noisy data makes it difficult to capture the true data distribution.

A straightforward approach to addressing this issue involves first denoising the data and then applying dataset distillation.
This includes identifying mislabeled data~\cite{han2018co, li2020dividemix, albert2022addressing, northcutt2021confident} and then either removing it or assigning a pseudo-label~\cite{tanaka2018joint, yi2019probabilistic}.
However, the independent denoising step has side effects in dataset distillation.
The denoising step, despite its effectiveness, may introduce biases by misidentifying noisy data, leading to the removal of useful samples or the assignment of incorrect labels. 
As a result, some erroneous labels or pseudo-labels inevitably propagate to the dataset distillation stage, amplifying errors and compromising its effectiveness.
This motivates us to design a trustworthy data-effective dataset distillation method through an iterative dual-loop optimization framework.

\section{Preliminary and Problem Setup}
\label{sec_3-1}

\begin{figure*}[ht]
    \centering
    \includegraphics[width=0.81\linewidth]{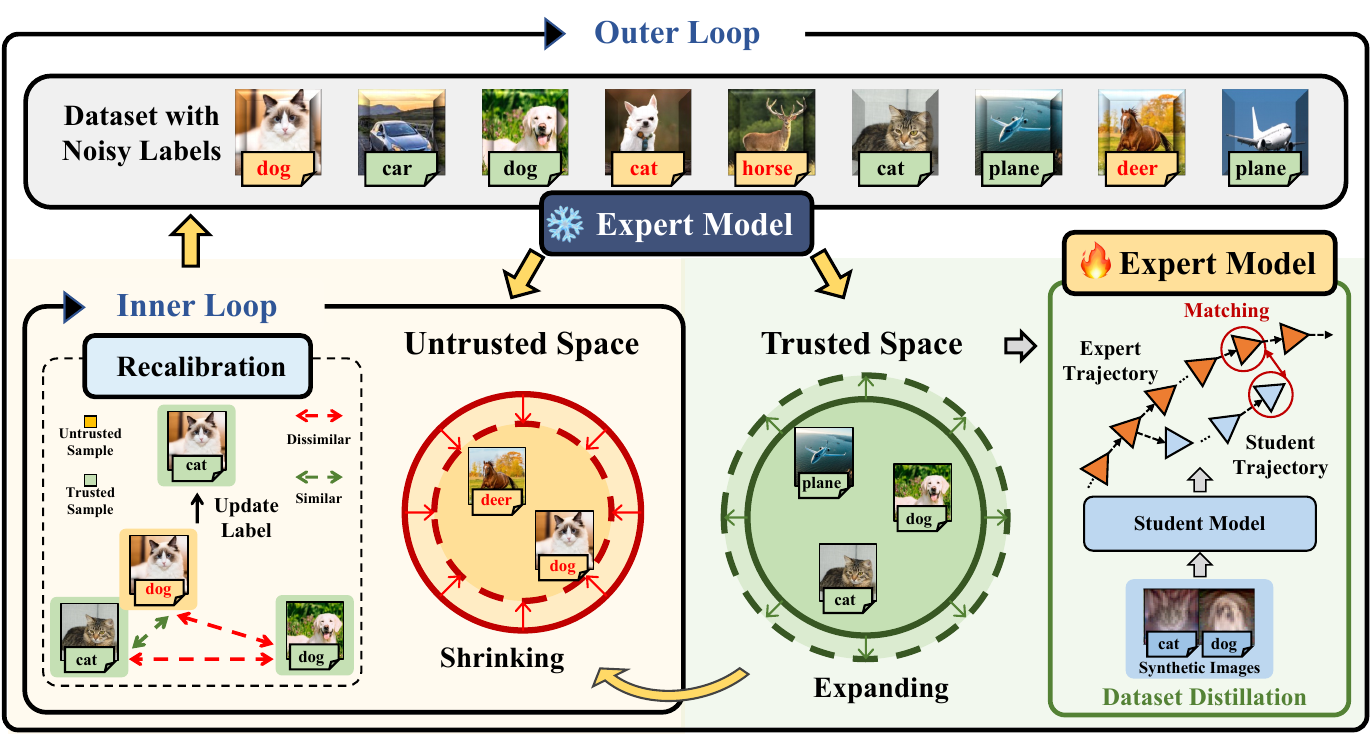}
    \vspace{-1em}
    \caption{
    Overview of the proposed Trust-Aware Diversion (TAD) dataset distillation method. 
    TAD introduces a dual-loop optimization framework for trustworthy dataset distillation. 
    The outer loop divides data into trusted and untrusted spaces, rerouting distillation toward reliable samples, while the inner loop refines untrusted samples for potential reuse. 
    Through iterative interaction, the two loops progressively expand the trusted space and mitigate the impact of noisy labels to achieve data-effective distillation.
    }
    \vspace{-1em}
    \label{fig3}
\end{figure*}

\noindent{\textbf{Dataset Distillation (DD).}} 
Given a large training set $\mathcal{T}=\left\{\left(\bm{x}_{i}, \bm{y}_{i}\right)\right\}_{i=1}^{|\mathcal{T}|}$ containing $|\mathcal{T}|$ images and their labels, dataset distillation aims to synthesize a smaller set $\mathcal{S}=\left\{\left(\bm{\widetilde x}_{i}, \bm{\widetilde y}_{i}\right)\right\}_{i=1}^{|\mathcal{S}|}$. 
The objective of dataset distillation is to optimize the synthetic set $\mathcal{S}$, making sure when a \emph{\textbf{student}} model $f_{\boldsymbol{\theta}^\mathcal{S}}$ is trained on the synthetic dataset $\mathcal{S}$, it can achieve comparable performance 
with an \emph{\textbf{expert}} model $f_{\boldsymbol{\theta}^\mathcal{T}}$ trained on the original dataset $\mathcal{T}$.

\noindent{\textbf{Trajectory Matching (TM).}} 
The expert model $f_{\boldsymbol{\theta}^\mathcal{T}}$ generates a sequence of parameters to make up an expert trajectory $\{\boldsymbol{\theta}^\mathcal{T}_i\}_{i=1}^{m}$. 
Similarly, the student model generates a trajectory $\{\boldsymbol{\theta}^\mathcal{S}_i\}_{i=1}^{n}$. 
TM-based methods perform distillation by matching the student trajectory with the expert trajectory. 
Starting from step $t$, the matching objective is defined as: 
\begin{equation}
    \label{eq1}
    \mathcal{L}_{TM} = \frac{\|\boldsymbol{\theta}^\mathcal{S}_{t+N} - \boldsymbol{\theta}^\mathcal{T}_{t+M}\|_2^2}
    {\|\boldsymbol{\theta}^\mathcal{T}_{t} - \boldsymbol{\theta}^\mathcal{T}_{t+M}\|_2^2}\,,
\end{equation}
where $N$ and $M$ are skipping steps ($N\ll M$). $\mathcal{L}_{TM}$ is adopted to optimize the synthetic set $\mathcal{S}$. 

\noindent{\textbf{Dataset Distillation with Noisy Labels (DDNL).}} 
Traditional dataset distillation assumes that labels in $\mathcal{T}$ are completely correct.
However, noisy labels are ubiquitous in real-world scenarios and impair model performance.
This work extends the dataset distillation to handle situations where the original training set $\mathcal{T}$ contains noisy data. 
Given $\mathcal{T}' = \left\{ \left( \bm{x}_i, \bm{y}'_i \right) \right\}_{i=1}^{|\mathcal{T}|}$, the label $\bm{y}'_i $ may be noisy.

To analyze the impact of noisy labels on dataset distillation, we investigate the behavior of cross-entropy (CE) loss and its gradient under noisy labels. 
CE is the most commonly used loss for trajectory matching in dataset distillation. Formally, CE loss and gradient are defined as: 
\begin{align}
\label{eq2}
& \mathcal{L}_\text{CE}\left(f_{\boldsymbol{\theta}}(\boldsymbol{x}), \boldsymbol{y}\right) =  -\sum_{k=1}^{K} \boldsymbol{y}_{(k)} \log (\sigma(\boldsymbol{z})_{k})\,, \\
\label{eq3}
& \frac{\partial \mathcal{L}_{\mathrm{CE}}(f_{\boldsymbol{\theta}}(\boldsymbol{x}), \boldsymbol{y})}{\partial \boldsymbol{\theta}} = \sum_{k=1}^{K} (\sigma(\boldsymbol{z})_k - \boldsymbol{y}_{(k)}) \frac{\partial \boldsymbol{z}_k}{\partial \boldsymbol{\theta}} \,,
\end{align}
where $\boldsymbol{z} = f_{\boldsymbol{\theta}}(\boldsymbol{x})$ denotes the predicted logit value for a sample $\boldsymbol{x}$ 
, $\sigma(\boldsymbol{z})_k$ denotes the $k$-th output of softmax function $\sigma$. 
$\mathcal{L}_{\mathrm{CE}}$ is unbounded and particularly vulnerable to noisy labels~\cite{ghosh2017robust, wei2023mitigating}.
If $\boldsymbol{y}$ is mis-labeled for a true class $k$, then $\boldsymbol{y}_{(k)} = 0$. But for a decent model, $\sigma(\boldsymbol{z})_k \rightarrow 1$. 
Consequently, the gradient updates in Eq.~\ref{eq3} will be misled by noisy labels, causing the parameters to gradually deviate from the ideal trajectory derived from clean data.

In dataset distillation, noisy labels will drive the expert trajectory $\{\boldsymbol{\theta}^\mathcal{T}_i\}_{i=1}^{m}$, away from the ideal path derived from clean data. 
For trajectory matching (Eq.~\ref{eq1}), the student model will align with this distorted expert trajectory and inherit the noise distortions, thus producing suboptimal synthetic data that ultimately degrade the overall performance of dataset distillation.

\section{Method}

To tackle the challenging DDNL problem, we propose a \textbf{Trust-Aware Diversion} (TAD) dataset distillation method, as shown in Fig.~\ref{fig3}.
Specifically, our proposed TAD introduces an end-to-end dual-loop (\ie, outer loop and inner loop) optimization framework for data-effective distillation. 
The outer loop (Sec.~\ref{sec:outer}) divides data into trusted and untrusted spaces, redirecting distillation toward trusted samples to minimize the impact of mislabeled samples on dataset distillation.
The inner loop (Sec.~\ref{sec:inner}) maximizes the distillation objective by recalibrating untrusted samples, thus transforming them into valuable ones for distillation.
This dual-loop iteratively refines and compensates for each other, gradually expanding the trusted space and shrinking the untrusted space.

\subsection{Outer Loop: Diverting Distillation for Trustworthy Learning}
 \label{sec:outer}
 
Effectively distinguishing clean samples from noisy ones is essential to fully leverage high-quality information while mitigating the adverse effects of noisy labels.
This distinction allows dataset distillation to prioritize learning from more reliable samples, ensuring a trustworthy distillation process.
During model training, the loss associated with each sample can help indicate whether it is a noisy sample~\cite{li2020dividemix, wei2020combating, yao2020searching, zhou2020robust}, as shown in Fig.~\ref{fig2}.
Clean samples are mutually consistent, allowing the model to produce gradient updates more efficiently and to train faster. 
In contrast, noisy samples often contain conflicting information, causing persistent inconsistencies in training progress.
Intuitively, samples with smaller loss values are highly likely to be clean samples during the training process.

Motivated by this, we interpret data as a mixture of two distinct sample categories: trusted and untrusted. 
To model this distinction, we represent loss distribution using a two-mode Gaussian Mixture Model (GMM), where one mode denotes trusted samples with lower losses, and the other represents untrusted samples with higher losses. 
This formulation is expressed as follows:
\begin{equation}
\begin{aligned}
p(\ell_i) = \sum_{k=1}^{K} \pi_k \cdot \mathcal{N}(\ell_i \mid \mu_k, \sigma_k^2),
\end{aligned}
\label{eq4}
\end{equation}
where $\ell_i= \mathcal{L}_{CE}(f_{\boldsymbol{\theta}}(\bm{x}_i), \bm{y}'_i)$ is the cross-entropy loss of the $i$-th sample and $\pi_k$ is the mixing coefficient.
For each sample, its trusted sample confidence $w_i$ is the posterior probability $p(k=1|\ell_i)$, where $k=1$ denotes the component of GMM with a smaller mean $\mu_k$ (\ie, smaller loss).
The GMM parameters are estimated with the Expectation-Maximization algorithm.
The trusted sample confidences $\{ w_i \}$ of the training dataset are then used to divide samples into trusted and untrusted spaces based on a threshold $\bm{\tau}$.

A straightforward way is to adopt a constant threshold (\eg, 0.5) for all samples, as in~\cite{arazo2019unsupervised, li2020dividemix, yang2022learning, zheltonozhskii2022contrast}. 
However, there are some issues: 
\begin{itemize} 
    \item \emph{
            The learning of the network is progressive. 
            In the early stages, confidence in identifying trusted samples is low.
            A fixed threshold cannot adapt to the estimation of trusted and untrusted samples throughout training.
    }
    \item \emph{
            Different classes have varying learning difficulty levels.
            Employing a unified threshold for all classes inadvertently causes class imbalance, particularly in datasets with numerous classes.
            }
\end{itemize}
To address the above issues, we propose a class-wise dynamic threshold to divide trusted samples and untrusted samples.
For a class $c$, we define the threshold $\bm{\tau}_c$ based on the posterior probabilities of GMM, \emph{i.e.}, the confidence $w_i$ of $\boldsymbol{x}_i$ being a trusted sample. 
Per-class mean confidence score is calculated as the threshold:
\begin{equation}
\begin{aligned}
\bm{\tau}_c = \frac{1}{N_c} \sum_{i=1}^{N_c} w_i, \; w_i = p(k=1|\ell_i) \; \text{ and } \; \bm{y}'_i = c,
\end{aligned}
\label{eq5}
\end{equation}
where $N_c$ is the number of samples in class $c$. 
The mean eliminates the effect of outliers, while the class-wise dynamic thresholds evolve as the training progresses to reflect the training dynamics and imbalanced class difficulties.

Building on existing dataset distillation approaches~\cite{cazenavette2022distillation, du2023minimizing, guo2024lossless, liu2024dataset}, we use CE loss to train expert networks solely on trusted space, preventing the model from learning misrepresented patterns in mislabeled data.
Additionally, to avoid overfitting to outliers and improve model consistency, we introduce a consistent regularization term $\mathcal{L}_{C}$ by reversing the roles of predicted probabilities and given labels.  
The training objective is formulated as follows:
\begin{equation}
\begin{aligned}
\mathcal{L} &= \mathcal{L}_{CE} + \lambda \mathcal{L}_{C} \\
&= - \sum_{i=1}^N \{\bm{y}'_i \log(f_{\boldsymbol{\theta}}(\bm{x}_i) + \lambda f_{\boldsymbol{\theta}}(\bm{x}_i) \log(\bm{y}'_i) \},
\end{aligned}
\label{eq6}
\end{equation}
where $\lambda$ is a coefficient to balance the two loss items.
For $\bm{y}'_i = 1$, the regularization term $\mathcal{L}_{C}=0$ does not contribute to the loss $\mathcal{L}$.
For $\bm{y}'_i = 0$, $\log(0)$ is defined as a negative constant to ensure numerical stability.
For the mislabeled samples, $\mathcal{L}_{CE}$ imposes a large loss, magnifying the effect of incorrect labels. 
This is counteracted by the $\mathcal{L}_{C}$ term, which reduces excessive loss and stabilizes the training process.

\subsection{Inner Loop: Trust-Aware Recalibration}
\label{sec:inner}

Although data has been partitioned into trusted and untrusted spaces in the outer loop, some noisy samples may still be mistakenly classified as trusted ones.
To address this issue, we introduce a reliability score in the inner loop to quantify the trustworthiness of each sample in the trusted space, providing a fine-grained evaluation of reliability.
Synthetic data aligning only the initial training phases of expert networks (\eg, the first four epochs) effectively capture representative patterns in real data~\cite{li2024prioritize, guo2024lossless}.
Thus, we use distilled data as stable anchors for assessing data quality.
A reliability score of each trusted sample is then calculated by computing its Mahalanobis distance~\cite{de2000mahalanobis} to the class distributions. 
For accurate label calibration of untrusted samples, we reference only the top-$k$ reliable trusted samples based on reliability scores.

First, anchors are defined as the synthetic images $\{\left(\hat{\bm{x}}_{i}, \hat{\bm{y}}_{i}\right)\}_{i=1}^{N_A}$ synthesized by only matching the early trajectory, which is feasible due to the memorization effects of deep networks~\cite{xia2020robust, arpit2017closer, liu2020early}.
Here $N_A = n \times C$, where $n$ represents the number of synthetic images per class, and $C$ is the total number of classes.
Second, the anchor images and trusted samples are embedded into the feature space by a feature extractor $f$ (\ie, ResNet~\cite{he2016deep}), which is pre-trained in a self-supervision manner (\eg, SimCLR~\cite{chen2020simple}, MoCo~\cite{he2020momentum}).
Third, we calculate the Mahalanobis distance~\cite{de2000mahalanobis} between trusted samples and the per-class anchor distribution. 
Mahalanobis distance is an effective multivariate distance metric used to calculate the distance between a point and a cluster, and it is widely used in multivariate anomaly detection, classification, and clustering analysis~\cite{colombo2022beyond, goswami2024fecam}. 
For each class $c$, the Mahalanobis distance is defined as:
\begin{equation}
\begin{aligned}
D_M(\bm{x}_{i}, \bm{\mu}_c) = \sqrt{(\bm{x}_{i} - \bm{\mu}_c)^T \mathbf{\Sigma}_c^{-1} (\bm{x}_{i} - \bm{\mu}_c)}, 
\end{aligned}
\label{eq7}
\end{equation}
where $\bm{x}_i$ denotes the $i$-th trusted sample, $\bm{\mu}_c$ and $\mathbf{\Sigma}_c^{-1}$ are the estimated class mean and covariance from anchors $\{\left(\hat{\bm{x}}_{i}, \hat{\bm{y}}_{i}\right)\}_{i=1}^{N_A}$ belonging to class $c$.
We then apply min-max normalization to re-scale each $D_M$ to the range $ \left [ 0,1 \right ]$. 
Then, we use the Mahalanobis distance to estimate the reliability of a trusted sample $\bm{x}_i$ and define $\bm{M}_i^c = D_M(\bm{x}_{i}, \bm{\mu}_c)$. 
$\bm{M}_i^c$ reflects the overall proximity of the sample to the anchor distribution of the class $c$.

To skip the unreliable trusted samples, we only refer to the top-$k$ reliable samples in the trusted space to calibrate the samples in the untrusted space. 
As Mahalanobis distance requires an accurate covariance structure, it is not suitable for comparing distances between untrusted and trusted samples. 
Instead, we use cosine similarity to compare trusted samples $\{\bm{x}_i\}$ with an untrusted sample $\bm{x}'_j$.
The calibrated pseudo label of $\bm{x}'_j$ for each class $c$ is computed as: 
\begin{equation}
p^c(\bm{x}'_j) = \sum_{i=1}^{k} \bm{M}_i^c \cdot \mathit{sim}(\bm{x}'_j, \bm{x}_i) \cdot \delta(y_i = c),
\label{eq10}
\end{equation}
where $\delta(y_i = c)$ is an indicator function equal to $1$ if the label $y_i$ of $\bm{x}_i$ is class $c$, and $0$ otherwise. 
The pseudo label for $\bm{x}'_j$ is then assigned based on the class $c$ with the highest score $p^c(\bm{x}'_j)$.

\section{Experiments}

\begin{table*}[ht!]
    \centering
    \renewcommand{\arraystretch}{.5}
    \setlength{\tabcolsep}{6pt}
    \caption{
    The test accuracy ($\%$) of dataset distillation methods with various ratios of symmetric (\textit{\textit{\textit{Sym.}}}) and asymmetric (\textit{Asym.}) noisy labels.
    Results (accuracy $\pm$ std) are shown for \textit{CIFAR-10}, \textit{CIFAR-100}, and \textit{Tiny ImageNet} datasets at different Image Per Class (IPC).
    }
    \tiny
\centering
\begin{tabular}{ccccccccccc}
\toprule
\multicolumn{2}{c|}{IPC}                                                           & \multicolumn{3}{c|}{1}                                                                                                     & \multicolumn{3}{c|}{10}                                                                                                    & \multicolumn{3}{c}{50}                                                                               \\ \midrule
\multicolumn{2}{c|}{Noise Ratio}                                                   & \multicolumn{1}{c|}{0\%}                           & \multicolumn{1}{c|}{20\%}         & \multicolumn{1}{c|}{40\%}         & \multicolumn{1}{c|}{0\%}                           & \multicolumn{1}{c|}{20\%}         & \multicolumn{1}{c}{40\%}         & \multicolumn{1}{c|}{0\%}                           & \multicolumn{1}{c|}{20\%}         & 40\%         \\ \midrule[.8pt] \midrule[.8pt]
\multicolumn{11}{c}{\textit{\textbf{CIFAR-10}}}                                                                                                                                                                                                                                                                                                                                                                                                     \\ \midrule[.8pt] \midrule[.8pt]
\multicolumn{1}{c|}{\multirow{20}{*}{\textit{Sym.}}}  & \multicolumn{1}{c|}{MTT~\cite{cazenavette2022distillation}}    & \multicolumn{1}{c|}{\multirow{3}{*}{46.3$\pm$0.8}} & \multicolumn{1}{c|}{45.2$\pm$0.3} & \multicolumn{1}{c|}{44.2$\pm$0.7} & \multicolumn{1}{c|}{\multirow{3}{*}{65.3$\pm$0.7}} & \multicolumn{1}{c|}{62.4$\pm$0.6} & \multicolumn{1}{c|}{57.5$\pm$1.2} & \multicolumn{1}{c|}{\multirow{3}{*}{71.6$\pm$0.2}} & \multicolumn{1}{c|}{68.6$\pm$0.5} & 60.2$\pm$0.7 \\ \cmidrule{2-2} \cmidrule{4-5} \cmidrule{7-8} \cmidrule{10-11}
\multicolumn{1}{c|}{}                       & \multicolumn{1}{c|}{\cellcolor[HTML]{EFEFEF} + TAD} & \multicolumn{1}{c|}{}                              & \multicolumn{1}{c|}{\cellcolor[HTML]{EFEFEF}46.2$\pm$0.4} & \multicolumn{1}{c|}{\cellcolor[HTML]{EFEFEF}46.1$\pm$0.6} & \multicolumn{1}{c|}{}                              & \multicolumn{1}{c|}{\cellcolor[HTML]{EFEFEF}64.7$\pm$0.5} & \multicolumn{1}{c|}{\cellcolor[HTML]{EFEFEF}64.6$\pm$0.6} & \multicolumn{1}{c|}{}                              & \multicolumn{1}{c|}{\cellcolor[HTML]{EFEFEF}71.5$\pm$0.2} & \cellcolor[HTML]{EFEFEF}71.2$\pm$0.3 \\ 
\cmidrule{2-11} 
\multicolumn{1}{c|}{}                       & \multicolumn{1}{c|}{FTD~\cite{du2023minimizing}}    & \multicolumn{1}{c|}{\multirow{3}{*}{46.8$\pm$0.3}} & \multicolumn{1}{c|}{45.3$\pm$1.0} & \multicolumn{1}{c|}{43.9$\pm$0.7} & \multicolumn{1}{c|}{\multirow{3}{*}{66.6$\pm$0.3}} & \multicolumn{1}{c|}{61.3$\pm$0.5} & \multicolumn{1}{c|}{56.2$\pm$0.9} & \multicolumn{1}{c|}{\multirow{3}{*}{73.8$\pm$0.2}} & \multicolumn{1}{c|}{67.9$\pm$0.7} & 61.1$\pm$0.6 \\ \cmidrule{2-2} \cmidrule{4-5} \cmidrule{7-8} \cmidrule{10-11} 
\multicolumn{1}{c|}{}                       & \multicolumn{1}{c|}{\cellcolor[HTML]{EFEFEF} + TAD} & \multicolumn{1}{c|}{}                              & \multicolumn{1}{c|}{\cellcolor[HTML]{EFEFEF}46.6$\pm$0.4} & \multicolumn{1}{c|}{\cellcolor[HTML]{EFEFEF}46.2$\pm$1.2} & \multicolumn{1}{c|}{}                              & \multicolumn{1}{c|}{\cellcolor[HTML]{EFEFEF}66.1$\pm$0.2} & \multicolumn{1}{c|}{\cellcolor[HTML]{EFEFEF}65.4$\pm$0.5} & \multicolumn{1}{c|}{}                              & \multicolumn{1}{c|}{\cellcolor[HTML]{EFEFEF}71.9$\pm$0.3} & \cellcolor[HTML]{EFEFEF}71.7$\pm$0.2 \\ \cmidrule{2-11}  
\multicolumn{1}{c|}{}                       & \multicolumn{1}{c|}{DATM~\cite{guo2024lossless}}   & \multicolumn{1}{c|}{\multirow{3}{*}{46.9$\pm$0.5}} & \multicolumn{1}{c|}{45.3$\pm$0.5} & \multicolumn{1}{c|}{42.9$\pm$0.3} & \multicolumn{1}{c|}{\multirow{3}{*}{66.8$\pm$0.2}} & \multicolumn{1}{c|}{62.1$\pm$0.1} & \multicolumn{1}{c|}{58.2$\pm$0.4} & \multicolumn{1}{c|}{\multirow{3}{*}{76.1$\pm$0.3}} & \multicolumn{1}{c|}{68.7$\pm$0.3} & 61.5$\pm$0.4 \\ \cmidrule{2-2} \cmidrule{4-5} \cmidrule{7-8} \cmidrule{10-11} 
\multicolumn{1}{c|}{}                       & \multicolumn{1}{c|}{\cellcolor[HTML]{EFEFEF} + TAD} & \multicolumn{1}{c|}{}                              & \multicolumn{1}{c|}{\cellcolor[HTML]{EFEFEF}46.1$\pm$0.9} & \multicolumn{1}{c|}{\cellcolor[HTML]{EFEFEF}45.8$\pm$0.2} & \multicolumn{1}{c|}{}                              & \multicolumn{1}{c|}{\cellcolor[HTML]{EFEFEF}66.0$\pm$0.3} & \multicolumn{1}{c|}{\cellcolor[HTML]{EFEFEF}65.5$\pm$0.6} & \multicolumn{1}{c|}{}                              & \multicolumn{1}{c|}{\cellcolor[HTML]{EFEFEF}72.6$\pm$0.2} & \cellcolor[HTML]{EFEFEF}71.6$\pm$0.2 \\ \cmidrule{2-11}  
\multicolumn{1}{c|}{}                       & \multicolumn{1}{c|}{ATT~\cite{liu2024dataset}}    & \multicolumn{1}{c|}{\multirow{3}{*}{48.3$\pm$1.0}} & \multicolumn{1}{c|}{44.5$\pm$0.9} & \multicolumn{1}{c|}{41.8$\pm$0.9} & \multicolumn{1}{c|}{\multirow{3}{*}{67.7$\pm$0.6}} & \multicolumn{1}{c|}{62.4$\pm$0.6} & \multicolumn{1}{c|}{58.1$\pm$0.6} & \multicolumn{1}{c|}{\multirow{3}{*}{74.5$\pm$0.4}} & \multicolumn{1}{c|}{69.3$\pm$0.2} & 63.8$\pm$0.3 \\ \cmidrule{2-2} \cmidrule{4-5} \cmidrule{7-8} \cmidrule{10-11} 
\multicolumn{1}{c|}{}                       & \multicolumn{1}{c|}{\cellcolor[HTML]{EFEFEF} + TAD} & \multicolumn{1}{c|}{}                              & \multicolumn{1}{c|}{\cellcolor[HTML]{EFEFEF}47.9$\pm$0.8} & \multicolumn{1}{c|}{\cellcolor[HTML]{EFEFEF}46.8$\pm$0.5} & \multicolumn{1}{c|}{}                              & \multicolumn{1}{c|}{\cellcolor[HTML]{EFEFEF}66.4$\pm$0.4} & \multicolumn{1}{c|}{\cellcolor[HTML]{EFEFEF}66.2$\pm$0.2} & \multicolumn{1}{c|}{}                              & \multicolumn{1}{c|}{\cellcolor[HTML]{EFEFEF}71.3$\pm$0.3} & \cellcolor[HTML]{EFEFEF}71.1$\pm$0.4 \\ \midrule
\multicolumn{1}{c|}{\multirow{20}{*}{\textit{Asym.}}} & \multicolumn{1}{c|}{MTT~\cite{cazenavette2022distillation}}    & \multicolumn{1}{c|}{\multirow{3}{*}{46.3$\pm$0.8}} & \multicolumn{1}{c|}{44.5$\pm$0.6} & \multicolumn{1}{c|}{40.9$\pm$1.7} & \multicolumn{1}{c|}{\multirow{3}{*}{65.3$\pm$0.7}} & \multicolumn{1}{c|}{58.7$\pm$0.3} & \multicolumn{1}{c|}{54.4$\pm$0.8} & \multicolumn{1}{c|}{\multirow{3}{*}{71.6$\pm$0.2}} & \multicolumn{1}{c|}{67.5$\pm$0.4} & 59.1$\pm$0.6 \\ \cmidrule{2-2} \cmidrule{4-5} \cmidrule{7-8} \cmidrule{10-11} 
\multicolumn{1}{c|}{}                       & \multicolumn{1}{c|}{\cellcolor[HTML]{EFEFEF} + TAD} & \multicolumn{1}{c|}{}                              & \multicolumn{1}{c|}{\cellcolor[HTML]{EFEFEF}45.8$\pm$0.9} & \multicolumn{1}{c|}{\cellcolor[HTML]{EFEFEF}45.6$\pm$1.4} & \multicolumn{1}{c|}{}                              & \multicolumn{1}{c|}{\cellcolor[HTML]{EFEFEF}64.2$\pm$0.3} & \multicolumn{1}{c|}{\cellcolor[HTML]{EFEFEF}63.9$\pm$0.3} & \multicolumn{1}{c|}{}                              & \multicolumn{1}{c|}{\cellcolor[HTML]{EFEFEF}71.1$\pm$0.2} & \cellcolor[HTML]{EFEFEF}70.0$\pm$0.3 \\ \cmidrule{2-11}  
\multicolumn{1}{c|}{}                       & \multicolumn{1}{c|}{FTD~\cite{du2023minimizing}}    & \multicolumn{1}{c|}{\multirow{3}{*}{46.8$\pm$0.3}} & \multicolumn{1}{c|}{45.0$\pm$0.6} & \multicolumn{1}{c|}{42.5$\pm$1.0} & \multicolumn{1}{c|}{\multirow{3}{*}{66.6$\pm$0.3}} & \multicolumn{1}{c|}{60.9$\pm$0.6} & \multicolumn{1}{c|}{53.4$\pm$0.6} & \multicolumn{1}{c|}{\multirow{3}{*}{73.8$\pm$0.2}} & \multicolumn{1}{c|}{63.6$\pm$0.7} & 53.3$\pm$0.8 \\ \cmidrule{2-2} \cmidrule{4-5} \cmidrule{7-8} \cmidrule{10-11} 
\multicolumn{1}{c|}{}                       & \multicolumn{1}{c|}{\cellcolor[HTML]{EFEFEF} + TAD} & \multicolumn{1}{c|}{}                              & \multicolumn{1}{c|}{\cellcolor[HTML]{EFEFEF}45.9$\pm$0.3} & \multicolumn{1}{c|}{\cellcolor[HTML]{EFEFEF}45.8$\pm$0.8} & \multicolumn{1}{c|}{}                              & \multicolumn{1}{c|}{\cellcolor[HTML]{EFEFEF}65.3$\pm$0.4} & \multicolumn{1}{c|}{\cellcolor[HTML]{EFEFEF}64.9$\pm$0.5} & \multicolumn{1}{c|}{}                              & \multicolumn{1}{c|}{\cellcolor[HTML]{EFEFEF}71.0$\pm$0.4} & \cellcolor[HTML]{EFEFEF}69.9$\pm$0.2 \\ \cmidrule{2-11}  
\multicolumn{1}{c|}{}                       & \multicolumn{1}{c|}{DATM~\cite{guo2024lossless}}   & \multicolumn{1}{c|}{\multirow{3}{*}{46.9$\pm$0.5}} & \multicolumn{1}{c|}{43.7$\pm$0.6} & \multicolumn{1}{c|}{38.7$\pm$0.3} & \multicolumn{1}{c|}{\multirow{3}{*}{66.8$\pm$0.2}} & \multicolumn{1}{c|}{61.8$\pm$0.4} & \multicolumn{1}{c|}{57.3$\pm$0.4} & \multicolumn{1}{c|}{\multirow{3}{*}{76.1$\pm$0.3}} & \multicolumn{1}{c|}{62.5$\pm$0.5} & 55.0$\pm$0.4 \\ \cmidrule{2-2} \cmidrule{4-5} \cmidrule{7-8} \cmidrule{10-11} 
\multicolumn{1}{c|}{}                       & \multicolumn{1}{c|}{\cellcolor[HTML]{EFEFEF} + TAD} & \multicolumn{1}{c|}{}                              & \multicolumn{1}{c|}{\cellcolor[HTML]{EFEFEF}46.0$\pm$0.3} & \multicolumn{1}{c|}{\cellcolor[HTML]{EFEFEF}45.4$\pm$0.6} & \multicolumn{1}{c|}{}                              & \multicolumn{1}{c|}{\cellcolor[HTML]{EFEFEF}64.8$\pm$0.5} & \multicolumn{1}{c|}{\cellcolor[HTML]{EFEFEF}64.3$\pm$0.3} & \multicolumn{1}{c|}{}                              & \multicolumn{1}{c|}{\cellcolor[HTML]{EFEFEF}72.4$\pm$0.2} & \cellcolor[HTML]{EFEFEF}70.7$\pm$0.3 \\ \cmidrule{2-11}  
\multicolumn{1}{c|}{}                       & \multicolumn{1}{c|}{ATT~\cite{liu2024dataset}}    & \multicolumn{1}{c|}{\multirow{3}{*}{48.3$\pm$1.0}} & \multicolumn{1}{c|}{43.5$\pm$0.7} & \multicolumn{1}{c|}{41.9$\pm$1.2} & \multicolumn{1}{c|}{\multirow{3}{*}{67.7$\pm$0.6}} & \multicolumn{1}{c|}{62.1$\pm$0.4} & \multicolumn{1}{c|}{54.6$\pm$0.6} & \multicolumn{1}{c|}{\multirow{3}{*}{74.5$\pm$0.4}} & \multicolumn{1}{c|}{67.3$\pm$0.5} & 61.5$\pm$0.2 \\ \cmidrule{2-2} \cmidrule{4-5} \cmidrule{7-8} \cmidrule{10-11} 
\multicolumn{1}{c|}{}                       & \multicolumn{1}{c|}{\cellcolor[HTML]{EFEFEF} + TAD} & \multicolumn{1}{c|}{}                              & \multicolumn{1}{c|}{\cellcolor[HTML]{EFEFEF}47.3$\pm$0.6} & \multicolumn{1}{c|}{\cellcolor[HTML]{EFEFEF}46.4$\pm$0.4} & \multicolumn{1}{c|}{}                              & \multicolumn{1}{c|}{\cellcolor[HTML]{EFEFEF}66.2$\pm$0.5} & \multicolumn{1}{c|}{\cellcolor[HTML]{EFEFEF}65.6$\pm$0.2} & \multicolumn{1}{c|}{}                              & \multicolumn{1}{c|}{\cellcolor[HTML]{EFEFEF}70.6$\pm$0.2} & \cellcolor[HTML]{EFEFEF}69.5$\pm$0.3 \\ \midrule
\multicolumn{2}{c|}{Full Dataset}                                                  & \multicolumn{9}{c}{84.8$\pm$0.1}                                                                                                                                                                                                                                                                                                                               \\ \midrule[.8pt] \midrule[.8pt]
\multicolumn{11}{c}{\textit{\textbf{CIFAR-100}}}                                                                                                                                                                                                                                                                                                                                                                                                    \\ \midrule[.8pt] \midrule[.8pt]
\multicolumn{1}{c|}{\multirow{20}{*}{\textit{Sym.}}}  & \multicolumn{1}{c|}{MTT~\cite{cazenavette2022distillation}}    & \multicolumn{1}{c|}{\multirow{3}{*}{24.3$\pm$0.3}} & \multicolumn{1}{c|}{23.3$\pm$0.4} & \multicolumn{1}{c|}{21.5$\pm$1.8} & \multicolumn{1}{c|}{\multirow{3}{*}{40.1$\pm$0.4}} & \multicolumn{1}{c|}{36.5$\pm$0.2} & \multicolumn{1}{c|}{31.4$\pm$0.4} & \multicolumn{1}{c|}{\multirow{3}{*}{47.7$\pm$0.2}} & \multicolumn{1}{c|}{43.8$\pm$0.4} & 37.2$\pm$0.4 \\ \cmidrule{2-2} \cmidrule{4-5} \cmidrule{7-8} \cmidrule{10-11} 
\multicolumn{1}{c|}{}                       & \multicolumn{1}{c|}{\cellcolor[HTML]{EFEFEF} + TAD} & \multicolumn{1}{c|}{}                              & \multicolumn{1}{c|}{\cellcolor[HTML]{EFEFEF}23.0$\pm$0.4} & \multicolumn{1}{c|}{\cellcolor[HTML]{EFEFEF}21.7$\pm$0.1} & \multicolumn{1}{c|}{}                              & \multicolumn{1}{c|}{\cellcolor[HTML]{EFEFEF}40.0$\pm$0.1} & \multicolumn{1}{c|}{\cellcolor[HTML]{EFEFEF}39.4$\pm$0.1} & \multicolumn{1}{c|}{}                              & \multicolumn{1}{c|}{\cellcolor[HTML]{EFEFEF}44.2$\pm$0.2} & \cellcolor[HTML]{EFEFEF}43.2$\pm$0.1 \\ \cmidrule{2-11}  
\multicolumn{1}{c|}{}                       & \multicolumn{1}{c|}{FTD~\cite{du2023minimizing}}    & \multicolumn{1}{c|}{\multirow{3}{*}{25.2$\pm$0.2}} & \multicolumn{1}{c|}{23.8$\pm$0.5} & \multicolumn{1}{c|}{21.8$\pm$0.5} & \multicolumn{1}{c|}{\multirow{3}{*}{43.4$\pm$0.3}} & \multicolumn{1}{c|}{38.9$\pm$0.5} & \multicolumn{1}{c|}{33.5$\pm$0.4} & \multicolumn{1}{c|}{\multirow{3}{*}{50.7$\pm$0.3}} & \multicolumn{1}{c|}{43.5$\pm$0.3} & 36.9$\pm$0.3 \\ \cmidrule{2-2} \cmidrule{4-5} \cmidrule{7-8} \cmidrule{10-11} 
\multicolumn{1}{c|}{}                       & \multicolumn{1}{c|}{\cellcolor[HTML]{EFEFEF} + TAD} & \multicolumn{1}{c|}{}                              & \multicolumn{1}{c|}{\cellcolor[HTML]{EFEFEF}24.1$\pm$0.2} & \multicolumn{1}{c|}{\cellcolor[HTML]{EFEFEF}22.3$\pm$0.2} & \multicolumn{1}{c|}{}                              & \multicolumn{1}{c|}{\cellcolor[HTML]{EFEFEF}40.4$\pm$0.4} & \multicolumn{1}{c|}{\cellcolor[HTML]{EFEFEF}39.4$\pm$0.1} & \multicolumn{1}{c|}{}                              & \multicolumn{1}{c|}{\cellcolor[HTML]{EFEFEF}45.6$\pm$0.1} & \cellcolor[HTML]{EFEFEF}44.0$\pm$0.1 \\ \cmidrule{2-11}  
\multicolumn{1}{c|}{}                       & \multicolumn{1}{c|}{DATM~\cite{guo2024lossless}}   & \multicolumn{1}{c|}{\multirow{3}{*}{27.9$\pm$0.2}} & \multicolumn{1}{c|}{23.1$\pm$0.2} & \multicolumn{1}{c|}{22.1$\pm$0.2} & \multicolumn{1}{c|}{\multirow{3}{*}{47.2$\pm$0.4}} & \multicolumn{1}{c|}{39.7$\pm$0.4} & \multicolumn{1}{c|}{37.1$\pm$0.4} & \multicolumn{1}{c|}{\multirow{3}{*}{55.0$\pm$0.2}} & \multicolumn{1}{c|}{44.3$\pm$0.2} & 40.2$\pm$0.2 \\ \cmidrule{2-2} \cmidrule{4-5} \cmidrule{7-8} \cmidrule{10-11} 
\multicolumn{1}{c|}{}                       & \multicolumn{1}{c|}{\cellcolor[HTML]{EFEFEF} + TAD} & \multicolumn{1}{c|}{}                              & \multicolumn{1}{c|}{\cellcolor[HTML]{EFEFEF}23.9$\pm$0.5} & \multicolumn{1}{c|}{\cellcolor[HTML]{EFEFEF}22.3$\pm$0.3} & \multicolumn{1}{c|}{}                              & \multicolumn{1}{c|}{\cellcolor[HTML]{EFEFEF}43.0$\pm$0.1} & \multicolumn{1}{c|}{\cellcolor[HTML]{EFEFEF}41.5$\pm$0.1} & \multicolumn{1}{c|}{}                              & \multicolumn{1}{c|}{\cellcolor[HTML]{EFEFEF}46.9$\pm$0.1} & \cellcolor[HTML]{EFEFEF}46.1$\pm$0.1 \\ \cmidrule{2-11}  
\multicolumn{1}{c|}{}                       & \multicolumn{1}{c|}{ATT~\cite{liu2024dataset}}    & \multicolumn{1}{c|}{\multirow{3}{*}{26.1$\pm$0.3}} & \multicolumn{1}{c|}{22.1$\pm$0.3} & \multicolumn{1}{c|}{19.9$\pm$0.8} & \multicolumn{1}{c|}{\multirow{3}{*}{44.2$\pm$0.5}} & \multicolumn{1}{c|}{37.3$\pm$0.4} & \multicolumn{1}{c|}{32.6$\pm$0.3} & \multicolumn{1}{c|}{\multirow{3}{*}{51.2$\pm$0.3}} & \multicolumn{1}{c|}{41.9$\pm$0.2} & 37.1$\pm$0.2 \\ \cmidrule{2-2} \cmidrule{4-5} \cmidrule{7-8} \cmidrule{10-11} 
\multicolumn{1}{c|}{}                       & \multicolumn{1}{c|}{\cellcolor[HTML]{EFEFEF} + TAD} & \multicolumn{1}{c|}{}                              & \multicolumn{1}{c|}{\cellcolor[HTML]{EFEFEF}22.9$\pm$0.2} & \multicolumn{1}{c|}{\cellcolor[HTML]{EFEFEF}22.9$\pm$0.4} & \multicolumn{1}{c|}{}                              & \multicolumn{1}{c|}{\cellcolor[HTML]{EFEFEF}42.8$\pm$0.1} & \multicolumn{1}{c|}{\cellcolor[HTML]{EFEFEF}41.5$\pm$0.2} & \multicolumn{1}{c|}{}                              & \multicolumn{1}{c|}{\cellcolor[HTML]{EFEFEF}45.5$\pm$0.1} & \cellcolor[HTML]{EFEFEF}44.2$\pm$0.2 \\ \midrule
\multicolumn{1}{c|}{\multirow{20}{*}{\textit{Asym.}}} & \multicolumn{1}{c|}{MTT~\cite{cazenavette2022distillation}}    & \multicolumn{1}{c|}{\multirow{3}{*}{24.3$\pm$0.3}} & \multicolumn{1}{c|}{22.5$\pm$0.4} & \multicolumn{1}{c|}{20.0$\pm$0.6} & \multicolumn{1}{c|}{\multirow{3}{*}{40.1$\pm$0.4}} & \multicolumn{1}{c|}{36.2$\pm$0.4} & \multicolumn{1}{c|}{30.9$\pm$0.3} & \multicolumn{1}{c|}{\multirow{3}{*}{47.7$\pm$0.2}} & \multicolumn{1}{c|}{43.3$\pm$0.2} & 36.1$\pm$0.4 \\ \cmidrule{2-2} \cmidrule{4-5} \cmidrule{7-8} \cmidrule{10-11} 
\multicolumn{1}{c|}{}                       & \multicolumn{1}{c|}{\cellcolor[HTML]{EFEFEF} + TAD} & \multicolumn{1}{c|}{}                              & \multicolumn{1}{c|}{\cellcolor[HTML]{EFEFEF}22.6$\pm$0.3} & \multicolumn{1}{c|}{\cellcolor[HTML]{EFEFEF}20.2$\pm$0.3} & \multicolumn{1}{c|}{}                              & \multicolumn{1}{c|}{\cellcolor[HTML]{EFEFEF}39.9$\pm$0.4} & \multicolumn{1}{c|}{\cellcolor[HTML]{EFEFEF}37.3$\pm$0.4} & \multicolumn{1}{c|}{}                              & \multicolumn{1}{c|}{\cellcolor[HTML]{EFEFEF}43.6$\pm$0.1} & \cellcolor[HTML]{EFEFEF}41.3$\pm$0.1 \\ \cmidrule{2-11}  
\multicolumn{1}{c|}{}                       & \multicolumn{1}{c|}{FTD~\cite{du2023minimizing}}    & \multicolumn{1}{c|}{\multirow{3}{*}{25.2$\pm$0.2}} & \multicolumn{1}{c|}{23.4$\pm$0.2} & \multicolumn{1}{c|}{21.0$\pm$0.5} & \multicolumn{1}{c|}{\multirow{3}{*}{43.4$\pm$0.3}} & \multicolumn{1}{c|}{38.4$\pm$0.4} & \multicolumn{1}{c|}{32.8$\pm$0.3} & \multicolumn{1}{c|}{\multirow{3}{*}{50.7$\pm$0.3}} & \multicolumn{1}{c|}{42.1$\pm$0.3} & 35.6$\pm$0.3 \\ \cmidrule{2-2} \cmidrule{4-5} \cmidrule{7-8} \cmidrule{10-11} 
\multicolumn{1}{c|}{}                       & \multicolumn{1}{c|}{\cellcolor[HTML]{EFEFEF} + TAD} & \multicolumn{1}{c|}{}                              & \multicolumn{1}{c|}{\cellcolor[HTML]{EFEFEF}23.2$\pm$0.2} & \multicolumn{1}{c|}{\cellcolor[HTML]{EFEFEF}21.7$\pm$0.1} & \multicolumn{1}{c|}{}                              & \multicolumn{1}{c|}{\cellcolor[HTML]{EFEFEF}39.6$\pm$0.5} & \multicolumn{1}{c|}{\cellcolor[HTML]{EFEFEF}37.4$\pm$0.2} & \multicolumn{1}{c|}{}                              & \multicolumn{1}{c|}{\cellcolor[HTML]{EFEFEF}44.2$\pm$0.1} & \cellcolor[HTML]{EFEFEF}41.5$\pm$0.2 \\ \cmidrule{2-11}  
\multicolumn{1}{c|}{}                       & \multicolumn{1}{c|}{DATM~\cite{guo2024lossless}}   & \multicolumn{1}{c|}{\multirow{3}{*}{27.9$\pm$0.2}} & \multicolumn{1}{c|}{22.9$\pm$0.1} & \multicolumn{1}{c|}{21.4$\pm$0.1} & \multicolumn{1}{c|}{\multirow{3}{*}{47.2$\pm$0.4}} & \multicolumn{1}{c|}{39.0$\pm$0.2} & \multicolumn{1}{c|}{36.3$\pm$0.4} & \multicolumn{1}{c|}{\multirow{3}{*}{55.0$\pm$0.2}} & \multicolumn{1}{c|}{43.5$\pm$0.1} & 38.4$\pm$0.1 \\ \cmidrule{2-2} \cmidrule{4-5} \cmidrule{7-8} \cmidrule{10-11} 
\multicolumn{1}{c|}{}                       & \multicolumn{1}{c|}{\cellcolor[HTML]{EFEFEF} + TAD} & \multicolumn{1}{c|}{}                              & \multicolumn{1}{c|}{\cellcolor[HTML]{EFEFEF}23.1$\pm$0.6} & \multicolumn{1}{c|}{\cellcolor[HTML]{EFEFEF}21.7$\pm$0.5} & \multicolumn{1}{c|}{}                              & \multicolumn{1}{c|}{\cellcolor[HTML]{EFEFEF}42.5$\pm$0.5} & \multicolumn{1}{c|}{\cellcolor[HTML]{EFEFEF}39.6$\pm$0.2} & \multicolumn{1}{c|}{}                              & \multicolumn{1}{c|}{\cellcolor[HTML]{EFEFEF}46.5$\pm$0.2} & \cellcolor[HTML]{EFEFEF}43.7$\pm$0.2 \\ \cmidrule{2-11}  
\multicolumn{1}{c|}{}                       & \multicolumn{1}{c|}{ATT~\cite{liu2024dataset}}    & \multicolumn{1}{c|}{\multirow{3}{*}{26.1$\pm$0.3}} & \multicolumn{1}{c|}{21.7$\pm$0.2} & \multicolumn{1}{c|}{19.4$\pm$0.4} & \multicolumn{1}{c|}{\multirow{3}{*}{44.2$\pm$0.5}} & \multicolumn{1}{c|}{36.7$\pm$0.5} & \multicolumn{1}{c|}{31.9$\pm$0.2} & \multicolumn{1}{c|}{\multirow{3}{*}{51.2$\pm$0.3}} & \multicolumn{1}{c|}{39.5$\pm$0.2} & 33.6$\pm$0.3 \\ \cmidrule{2-2} \cmidrule{4-5} \cmidrule{7-8} \cmidrule{10-11} 
\multicolumn{1}{c|}{}                       & \multicolumn{1}{c|}{\cellcolor[HTML]{EFEFEF} + TAD} & \multicolumn{1}{c|}{}                              & \multicolumn{1}{c|}{\cellcolor[HTML]{EFEFEF}23.3$\pm$0.4} & \multicolumn{1}{c|}{\cellcolor[HTML]{EFEFEF}20.3$\pm$0.3} & \multicolumn{1}{c|}{}                              & \multicolumn{1}{c|}{\cellcolor[HTML]{EFEFEF}41.7$\pm$0.4} & \multicolumn{1}{c|}{\cellcolor[HTML]{EFEFEF}39.5$\pm$0.1} & \multicolumn{1}{c|}{}                              & \multicolumn{1}{c|}{\cellcolor[HTML]{EFEFEF}44.5$\pm$0.3} & \cellcolor[HTML]{EFEFEF}42.5$\pm$0.3 \\ \midrule
\multicolumn{2}{c|}{Full Dataset}                                                  & \multicolumn{9}{c}{56.2$\pm$0.3}                                                                                                                                                                                                                                                                                                                               \\ \midrule[.8pt] \midrule[.8pt]
\multicolumn{11}{c}{\textit{\textbf{Tiny ImageNet}}}                                                                                                                                                                                                                                                                                                                                                                                                \\ \midrule[.8pt] \midrule[.8pt]
\multicolumn{1}{c|}{\multirow{20}{*}{\textit{Sym.}}}  & \multicolumn{1}{c|}{MTT~\cite{cazenavette2022distillation}}    & \multicolumn{1}{c|}{\multirow{3}{*}{8.8$\pm$0.3}}  & \multicolumn{1}{c|}{8.0$\pm$0.3}  & \multicolumn{1}{c|}{7.0$\pm$0.4}  & \multicolumn{1}{c|}{\multirow{3}{*}{23.2$\pm$0.2}} & \multicolumn{1}{c|}{18.5$\pm$0.2} & \multicolumn{1}{c|}{16.3$\pm$0.3} & \multicolumn{1}{c|}{\multirow{3}{*}{28.0$\pm$0.3}} & \multicolumn{1}{c|}{22.8$\pm$0.2} & 16.8$\pm$0.2 \\ \cmidrule{2-2} \cmidrule{4-5} \cmidrule{7-8} \cmidrule{10-11} 
\multicolumn{1}{c|}{}                       & \multicolumn{1}{c|}{\cellcolor[HTML]{EFEFEF} + TAD}         & \multicolumn{1}{c|}{}                              & \multicolumn{1}{c|}{\cellcolor[HTML]{EFEFEF}8.2$\pm$0.1}  & \multicolumn{1}{c|}{\cellcolor[HTML]{EFEFEF}7.5$\pm$0.5}  & \multicolumn{1}{c|}{}                              & \multicolumn{1}{c|}{\cellcolor[HTML]{EFEFEF}22.3$\pm$0.1} & \multicolumn{1}{c|}{\cellcolor[HTML]{EFEFEF}21.9$\pm$0.5} & \multicolumn{1}{c|}{}                              & \multicolumn{1}{c|}{\cellcolor[HTML]{EFEFEF}27.5$\pm$0.2} & \cellcolor[HTML]{EFEFEF}26.7$\pm$0.2 \\ \cmidrule{2-11}  
\multicolumn{1}{c|}{}                       & \multicolumn{1}{c|}{FTD~\cite{du2023minimizing}}             & \multicolumn{1}{c|}{\multirow{3}{*}{10.4$\pm$0.3}} & \multicolumn{1}{c|}{9.1$\pm$0.4}  & \multicolumn{1}{c|}{8.4$\pm$0.3}  & \multicolumn{1}{c|}{\multirow{3}{*}{24.5$\pm$0.2}} & \multicolumn{1}{c|}{20.2$\pm$0.2} & \multicolumn{1}{c|}{16.4$\pm$0.5} & \multicolumn{1}{c|}{\multirow{3}{*}{\textbackslash}}            & \multicolumn{1}{c|}{\textbackslash}            & \textbackslash            \\ \cmidrule{2-2} \cmidrule{4-5} \cmidrule{7-8} \cmidrule{10-11} 
\multicolumn{1}{c|}{}                       & \multicolumn{1}{c|}{\cellcolor[HTML]{EFEFEF} + TAD}          & \multicolumn{1}{c|}{}                              & \multicolumn{1}{c|}{\cellcolor[HTML]{EFEFEF}10.4$\pm$0.4} & \multicolumn{1}{c|}{\cellcolor[HTML]{EFEFEF}10.2$\pm$0.6} & \multicolumn{1}{c|}{}                              & \multicolumn{1}{c|}{\cellcolor[HTML]{EFEFEF}24.3$\pm$0.4}             & \multicolumn{1}{c|}{\cellcolor[HTML]{EFEFEF}23.6$\pm$0.3} & \multicolumn{1}{c|}{}                              & \multicolumn{1}{c|}{\textbackslash}            & \textbackslash            \\ \cmidrule{2-11}  
\multicolumn{1}{c|}{}                       & \multicolumn{1}{c|}{DATM~\cite{guo2024lossless}}            & \multicolumn{1}{c|}{\multirow{3}{*}{17.1$\pm$0.3}} & \multicolumn{1}{c|}{10.4$\pm$0.3} & \multicolumn{1}{c|}{9.4$\pm$0.5}  & \multicolumn{1}{c|}{\multirow{3}{*}{31.1$\pm$0.3}} & \multicolumn{1}{c|}{24.6$\pm$0.3} & \multicolumn{1}{c|}{22.6$\pm$0.3} & \multicolumn{1}{c|}{\multirow{3}{*}{39.7$\pm$0.3}} & \multicolumn{1}{c|}{30.3$\pm$0.4} & 29.2$\pm$0.4 \\ \cmidrule{2-2} \cmidrule{4-5} \cmidrule{7-8} \cmidrule{10-11} 
\multicolumn{1}{c|}{}                       & \multicolumn{1}{c|}{\cellcolor[HTML]{EFEFEF} + TAD}          & \multicolumn{1}{c|}{}                              & \multicolumn{1}{c|}{\cellcolor[HTML]{EFEFEF}14.2$\pm$0.1}             & \multicolumn{1}{c|}{\cellcolor[HTML]{EFEFEF}13.9$\pm$0.1} & \multicolumn{1}{c|}{}                              & \multicolumn{1}{c|}{\cellcolor[HTML]{EFEFEF}27.2$\pm$0.3}             & \multicolumn{1}{c|}{\cellcolor[HTML]{EFEFEF}25.6$\pm$0.1} & \multicolumn{1}{c|}{}                              & \multicolumn{1}{c|}{\cellcolor[HTML]{EFEFEF}33.9$\pm$0.1}             & \cellcolor[HTML]{EFEFEF}32.1$\pm$0.1 \\ \cmidrule{2-11}  
\multicolumn{1}{c|}{}                       & \multicolumn{1}{c|}{ATT~\cite{liu2024dataset}}             & \multicolumn{1}{c|}{\multirow{3}{*}{11.0$\pm$0.5}} & \multicolumn{1}{c|}{7.1$\pm$0.3}  & \multicolumn{1}{c|}{6.6$\pm$0.1}  & \multicolumn{1}{c|}{\multirow{3}{*}{25.8$\pm$0.4}} & \multicolumn{1}{c|}{20.1$\pm$0.2} & \multicolumn{1}{c|}{16.9$\pm$0.2} & \multicolumn{1}{c|}{\multirow{3}{*}{\textbackslash}}            & \multicolumn{1}{c|}{\textbackslash}            & \textbackslash            \\ \cmidrule{2-2} \cmidrule{4-5} \cmidrule{7-8} \cmidrule{10-11} 
\multicolumn{1}{c|}{}                       & \multicolumn{1}{c|}{\cellcolor[HTML]{EFEFEF} + TAD}        & \multicolumn{1}{c|}{}                              & \multicolumn{1}{c|}{\cellcolor[HTML]{EFEFEF}10.2$\pm$0.2}             & \multicolumn{1}{c|}{\cellcolor[HTML]{EFEFEF}9.4$\pm$0.1}  & \multicolumn{1}{c|}{}                              & \multicolumn{1}{c|}{\cellcolor[HTML]{EFEFEF}24.9$\pm$0.2}             & \multicolumn{1}{c|}{\cellcolor[HTML]{EFEFEF}23.8$\pm$0.5}             & \multicolumn{1}{c|}{}                              & \multicolumn{1}{c|}{\textbackslash}            & \textbackslash            \\ \midrule
\multicolumn{2}{c|}{Full Dataset}                                                  & \multicolumn{9}{c}{37.6$\pm$0.4}                                                                                                                                                                                                                                                                                                                               \\ \bottomrule
\end{tabular}
\label{tab1}
\end{table*}

\subsection{Datasets}
\noindent{\textbf{Generation of Simulated Datasets.}} 
Our experiments are conducted on CIFAR-10, CIFAR-100, and Tiny ImageNet with two types of synthetic noise (\ie, symmetric and asymmetric).
\textit{Symmetric noise} is generated by uniformly flipping labels in each class to incorrect labels from other classes.
\textit{Asymmetric noise} flips the labels within a specific set of classes.
For CIFAR-10, labels are flipped as follows: TRUCK $\rightarrow$ AUTOMOBILE, BIRD $\rightarrow$ AIRPLANE, DEER $\rightarrow$ HORSE, CAT $\rightarrow$ DOG. 
For CIFAR-100, the 100 classes are grouped into 20 super-classes, and each has 5 sub-classes. 
Each class is then flipped to the next within the same super-class.
We use noise ratios of $20\%$ and $40\%$ in our experiments to evaluate the effectiveness of our proposed method.

\noindent{\textbf{Real-World Datasets.}}
We validate our method on the real-world datasets CIFAR-10N and CIFAR-100N, which contain human-annotated noisy labels obtained through Amazon Mechanical Turk~\cite{wei2022learning}.
Our evaluation primarily considers the CIFAR-10N (Worst) and CIFAR-100N (Fine) label sets, with noise ratios of 40.21$\%$ and 40.20$\%$, respectively.

\subsection{Experimental Setup}

Our method builds upon trajectory matching, and we conduct comprehensive comparisons with several baseline approaches, including MTT~\cite{cazenavette2022distillation}, FTD~\cite{du2023minimizing}, DATM~\cite{guo2024lossless}, and ATT~\cite{liu2024dataset}.
The detailed configurations of hyperparameters in the buffer phase and distillation phase align with the settings described in the corresponding works.
Experimental results for additional distillation methods (\eg, gradient matching, distribution matching, \etc) can be found in the appendix.
Without explicit mention, we use ConvNet by default to conduct experiments.
Following previous methods~\cite{cazenavette2022distillation, du2023minimizing, guo2024lossless, liu2024dataset}, we use a 3-layer ConvNet for CIFAR-10 and CIFAR-100, and a 4-layer ConvNet for Tiny ImageNet.

\subsection{Results}

\begin{table}[ht]
    \centering
    \renewcommand{\arraystretch}{1.05}
    \setlength{\tabcolsep}{5pt}
    \caption{
    Comparison of dataset distillation methods on \textit{CIFAR-10N (Worst)} and \textit{CIFAR-100N (Fine)} label sets. 
    Performance results (accuracy ± standard deviation) are shown across different Images Per Class (IPC).
    }
    \tiny
\begin{tabular}{c|c|c|c|c}
\toprule	
Dataset / IPC & Method & 1 & 10 & 50 \\ \midrule[.8pt] \midrule[.8pt]
\multirow{7}{*}{CIFAR-10N} & MTT~\cite{cazenavette2022distillation}  &   43.8$\pm$0.5 & 55.7$\pm$0.5 & 62.1$\pm$0.4 \\ 
& \cellcolor[HTML]{EFEFEF}+ TAD & \cellcolor[HTML]{EFEFEF}46.6$\pm$0.2  &\cellcolor[HTML]{EFEFEF}64.4$\pm$0.5  & \cellcolor[HTML]{EFEFEF}69.9$\pm$0.2 \\  
\cmidrule{2-5}
&FTD~\cite{du2023minimizing}               &   43.4$\pm$0.6 & 55.4$\pm$0.6 & 60.3$\pm$0.8 \\ 
& \cellcolor[HTML]{EFEFEF}+ TAD   &   \cellcolor[HTML]{EFEFEF}45.6$\pm$0.4 & \cellcolor[HTML]{EFEFEF}63.8$\pm$0.2 & \cellcolor[HTML]{EFEFEF}69.2$\pm$0.3  \\ \cmidrule{2-5}
&ATT~\cite{liu2024dataset}                 &   42.3$\pm$0.7 & 58.3$\pm$0.4 & 61.9$\pm$0.2 \\ 
&  \cellcolor[HTML]{EFEFEF}+ TAD                                    & \cellcolor[HTML]{EFEFEF}45.9$\pm$0.4   & \cellcolor[HTML]{EFEFEF}65.7$\pm$0.2 & \cellcolor[HTML]{EFEFEF}68.9$\pm$0.2 \\ \midrule[.8pt] \midrule[.8pt]
\multirow{7}{*}{CIFAR-100N}& MTT~\cite{cazenavette2022distillation}  &   20.2$\pm$0.4 & 32.8$\pm$0.3 & 38.2$\pm$0.3 \\ 
&   \cellcolor[HTML]{EFEFEF}+ TAD                                 &   \cellcolor[HTML]{EFEFEF}20.1$\pm$0.3 & \cellcolor[HTML]{EFEFEF}36.4$\pm$0.2 & \cellcolor[HTML]{EFEFEF} 42.1$\pm$0.1 \\ \cmidrule{2-5}
& FTD~\cite{du2023minimizing}            &     20.3$\pm$0.3 & 32.7$\pm$0.2 & 37.8$\pm$0.2 \\ 
& \cellcolor[HTML]{EFEFEF}+ TAD                                 &   \cellcolor[HTML]{EFEFEF}20.7$\pm$0.3 & \cellcolor[HTML]{EFEFEF}35.1$\pm$0.2 & \cellcolor[HTML]{EFEFEF}41.8$\pm$0.1\\ \cmidrule{2-5}
& ATT~\cite{liu2024dataset}              &    20.1$\pm$0.4 & 32.0$\pm$0.4 & 36.3$\pm$0.2\\ 
& \cellcolor[HTML]{EFEFEF}+ TAD                                 &  \cellcolor[HTML]{EFEFEF}21.4$\pm$0.2    &\cellcolor[HTML]{EFEFEF}35.9$\pm$0.1  & \cellcolor[HTML]{EFEFEF}41.2$\pm$0.3 \\ 
\bottomrule
\end{tabular}
\label{tab2}
\end{table}

\begin{table}[ht]
    \centering
    \renewcommand{\arraystretch}{1.05}
    \setlength{\tabcolsep}{6pt}
    \tiny
    \caption{
        The ablation study examines the effectiveness of different components, including consistent regularization (\textit{Reg.}), Outer Loop (\textit{OL.}), and inner loop (\textit{IL.}). 
        Each row represents a different combination of these components with checkmarks. 
        The baseline is DATM~\cite{guo2024lossless}.
        Experiments are conducted on CIFAR-10 with asymmetric noise ratios of 20\% and 40\%.
        IPC is set to 50.
    }
    \scriptsize
\begin{tabular}{c|ccc|cc}
\toprule	
Config            & \textit{Reg.}             & \textit{OL.}             & \textit{IL.}           & \textit{Asym.} 20\%         & \textit{Asym.} 40\%         \\ \midrule \midrule
A &                           &                           &                           & 62.5$\pm$0.5 & 55.0$\pm$0.4 \\ \midrule 
B                  & \checkmark &                           &                           &              64.1$\pm$0.3&              58.8$\pm$0.5\\ \midrule
C                  &                           & \checkmark &                           &              69.4$\pm$0.5&              65.1$\pm$0.5\\ \midrule 
D                  & \checkmark & \checkmark &                           &              70.7$\pm$0.2&              67.3$\pm$0.3\\ \midrule 
E                  &                           & \checkmark & \checkmark &              69.3$\pm$0.5&              67.6$\pm$0.4\\ \midrule 
F                  & \checkmark & \checkmark & \checkmark & 72.4$\pm$0.2 & 70.7$\pm$0.3 \\
\bottomrule
\end{tabular}
\label{tab3}
\end{table}

\noindent{\textbf{Dataset Distillation on Simulated Datasets.}} 
We conduct comprehensive experiments to compare the performance of previous dataset distillation methods with noisy labels.
In these experiments, we consider the effects of different noise types and noise ratios on various trajectory matching-based methods at different Images Per Class (IPC) configurations. 
This evaluation is performed on multiple datasets to ensure a thorough and reliable comparison.
As shown in Table~\ref{tab1}, noisy labels lead to a maximum accuracy drop of over 20\% in dataset distillation (\eg, for CIFAR-10 with 40\% asymmetric noise, FTD and DATM show significant accuracy drops when the IPC is set to 50).
In general, dataset distillation methods exhibit poorer performance with asymmetric noisy labels compared to symmetric ones. 
This is because asymmetric noise flips labels to closely related classes, making it more challenging for the model to distinguish between correct and noisy labels. 
Additionally, as IPC increases, the influence of noisy labels on dataset distillation grows increasingly significant.
Higher IPC requires the student model to align with the later stages of the training trajectory, where noisy labels exert a greater influence.
Our method iteratively refines samples to minimize the impact of noise on training trajectories. 
As a result, our approach achieves significant performance improvements.

In our evaluation of CIFAR-10, CIFAR-100, and Tiny ImageNet, we observe that our method achieves the most noticeable improvement on CIFAR-10.
This is because CIFAR-10 has fewer classes (10) and is relatively easy. 
In contrast, CIFAR-100 (100 classes) and Tiny ImageNet (200 classes) have more classes, which increases the difficulty and complexity. 
Our method demonstrates significant improvements on CIFAR-100 and Tiny ImageNet.
Overall, our method adapts effectively to various levels of dataset complexity and noise, demonstrating its robustness and versatility.
More results can be found in the appendix.

\begin{table}[ht]
    \centering
    \renewcommand{\arraystretch}{1.05}
    \setlength{\tabcolsep}{5pt}
    \caption{
    Comparison with learning from noisy labels methods. 
    Experiments are conducted on CIFAR-100 with symmetric noise.
    }
    \tiny
\begin{tabular}{c|cc|cc}
\toprule	
IPC           & \multicolumn{2}{c|}{10}               & \multicolumn{2}{c}{50}      \\ \midrule
Noise Ratio   & \multicolumn{1}{c}{\textit{Sym.} 20\%} & \textit{Sym.} 40\% & \multicolumn{1}{c}{\textit{Sym.} 20\%} & \textit{Sym.} 40\% \\ \midrule[.8pt] \midrule[.8pt]
MTT~\cite{cazenavette2022distillation}      & \multicolumn{1}{c}{36.5$\pm$0.2}   &    31.4$\pm$0.4& \multicolumn{1}{c}{43.8$\pm$0.4}   &    37.2$\pm$0.4\\ \midrule
+ C2D~\cite{zheltonozhskii2022contrast}     & \multicolumn{1}{c}{36.3$\pm$0.1}   &    35.6$\pm$0.5& \multicolumn{1}{c}{42.6$\pm$0.5}   &    40.6$\pm$0.6\\ \midrule
+ L2B-DivideMix~\cite{zhou2024l2b}          & \multicolumn{1}{c}{37.1$\pm$0.3}   &    36.0$\pm$0.4& \multicolumn{1}{c}{43.9$\pm$0.4}   &    41.4$\pm$0.5\\ \midrule
\rowcolor[HTML]{EFEFEF} + TAD                                        & \multicolumn{1}{c}{40.0$\pm$0.1}   &    39.4$\pm$0.1& \multicolumn{1}{c}{44.2$\pm$0.2}   &    43.2$\pm$0.1\\ \midrule[.8pt] \midrule[.8pt]
FTD~\cite{du2023minimizing}                 & \multicolumn{1}{c}{38.9$\pm$0.5}   &    33.5$\pm$0.4& \multicolumn{1}{c}{43.5$\pm$0.3}   &    36.9$\pm$0.3\\ \midrule
+ C2D~\cite{zheltonozhskii2022contrast}     & \multicolumn{1}{c}{38.2$\pm$0.2  }   &    36.7$\pm$0.3& \multicolumn{1}{c}{42.4$\pm$0.2}   &    41.7$\pm$0.3\\ \midrule
+ L2B-DivideMix~\cite{zhou2024l2b}          & \multicolumn{1}{c}{39.0$\pm$0.5}   &    38.1$\pm$0.4& \multicolumn{1}{c}{44.3$\pm$0.3}   &    42.2$\pm$0.4\\ \midrule
\rowcolor[HTML]{EFEFEF} + TAD                                        & \multicolumn{1}{c}{40.4$\pm$0.4}   &    39.4$\pm$0.1& \multicolumn{1}{c}{45.6$\pm$0.1}   &    44.0$\pm$0.1\\ \midrule[.8pt] \midrule[.8pt]
ATT~\cite{liu2024dataset}                   & \multicolumn{1}{c}{37.3$\pm$0.4}   &    32.6$\pm$0.3& \multicolumn{1}{c}{41.9$\pm$0.2}   &    37.1$\pm$0.2\\ \midrule
+ C2D~\cite{zheltonozhskii2022contrast}     & \multicolumn{1}{c}{36.9$\pm$0.4}   &35.8$\pm$0.2    & \multicolumn{1}{c}{40.8$\pm$0.3}   & 39.8$\pm$0.3    \\ \midrule
+ L2B-DivideMix~\cite{zhou2024l2b}          & \multicolumn{1}{c}{38.2$\pm$0.2}   &36.5$\pm$0.4    & \multicolumn{1}{c}{43.1$\pm$0.2}   & 41.0$\pm$0.3    \\ \midrule
\rowcolor[HTML]{EFEFEF} + TAD                                        & \multicolumn{1}{c}{42.8$\pm$0.1}   &    41.5$\pm$0.2& \multicolumn{1}{c}{45.5$\pm$0.1}   &    44.2$\pm$0.2\\ 
\bottomrule
\end{tabular}
\label{tab4}
\end{table}

\noindent{\textbf{Dataset Distillation on Real-world Datasets.}} 
To further verify the effectiveness of our method in real-world scenarios with noisy labels, we conduct comparative experiments on CIFAR-10N (Worst) and CIFAR-100N (Fine).
The experiments compare different dataset distillation methods, including MTT~\cite{cazenavette2022distillation}, FTD~\cite{du2023minimizing}, and ATT~\cite{liu2024dataset}, with and without our proposed method.
As shown in Table~\ref{tab2}, our proposed method (\ie, + Ours) generally enhances performance across different distillation methods on both CIFAR-10N (Worst) and CIFAR-100N (Fine) datasets.
On CIFAR-10N (Worst), our method significantly improves the accuracy of all baseline methods, especially as IPC increases (\eg, IPC=10 and 50). 
For instance, the accuracy of FTD improves from 60.3\% to 69.2\% at IPC=50, demonstrating notable enhancement.
On CIFAR-100N (Fine), the improvement is consistent, especially at higher IPC. 
Overall, our method demonstrates robustness and effectiveness in addressing realistic noisy labels.

\subsection{Ablation Study}
We conduct ablation studies to validate the effectiveness of our components.
The baseline for comparison is DATM~\cite{guo2024lossless}. 
IPC is set to 50, and evaluations are conducted on CIFAR-10 with asymmetric noise ratios of 20\% and 40\%.
As can be seen in Table~\ref{tab3}, the consistent regularization term effectively mitigates overfitting to noisy samples, improving performance (\eg, Config B, D, and F), particularly at higher noise ratios.
The outer loop (\ie, Config C) effectively filters out noisy samples. 
This keeps the expert model focused on accurate representations, minimizing the influence of noise and enabling high-quality distillation.
For instance, at a noise ratio of 40\%, the outer loop improves the performance of dataset distillation from  55.0\% to 65.1\%.
Additionally, the inner loop effectively corrects noisy labels by leveraging reliable anchors, allowing more data to contribute to the training process.
The collaboration of these modules proves effective in addressing dataset distillation with noisy labels, achieving superior performance.

\subsection{Comparison with Learning from Noisy Labels}

To further validate our method, we benchmark it against state-of-the-art LNL (Learning with Noisy Labels) methods, C2D~\cite{zheltonozhskii2022contrast} and L2B-DivideMix~\cite{zhou2024l2b}.
However, we find that using LNL methods to train the expert model creates a challenge. 
The student model struggles to align with the expert trajectory.
We speculate that LNL methods rely on semi-supervised learning, while the student model depends solely on CE. 
This difference causes a divergence in their optimization.
To address this, we use a straightforward approach.
We apply LNL methods to generate pseudo labels for the training set. 
Then, we perform dataset distillation directly using these pseudo-labeled data.

For the C2D, the pre-training model uses ResNet50 optimized with SimCLR~\cite{chen2020simple}.
Experiments are conducted on CIFAR-100 with symmetric noise.
As shown in Table~\ref{tab4}, our proposed method outperforms LNL methods.
Notably, at a noise ratio of 20\%, LNL methods provide minimal improvement to dataset distillation performance and may even degrade it (\eg, C2D).
The reason is that LNL methods cannot assign correct labels to all samples, and noisy samples greatly affect the expert trajectory.
Our proposed method, in contrast, is conservative. 
We model the per-sample loss to distinguish clean and noisy samples.
This approach helps us keep a training set with a high proportion of clean samples.
Training the expert models exclusively on this clean subset significantly reduces the impact of noisy labels on their training.

\section{Conclusion}

In this work, we propose a Trust-Aware Diversion (TAD) dataset distillation method to tackle the realistic yet challenging task of Dataset Distillation with Noisy Labels (DDNL). 
TAD introduces an iterative dual-loop (\ie, outer loop and inner loop) optimization framework to effectively mitigate the negative impact of mislabeled samples.
The outer loop partitions data into trusted and untrusted spaces, ensuring that distillation is guided by reliable samples, while the inner loop recalibrates untrusted samples, transforming them into valuable ones.
Through this iterative refinement, TAD expands the trusted space while shrinking the untrusted space, leading to more robust and reliable dataset distillation.
Extensive experiments on three benchmark datasets (CIFAR-10, CIFAR-100, and Tiny ImageNet) demonstrate that TAD achieves significant improvements under three noisy labeling settings: symmetric, asymmetric, and real-world noise. 
These results highlight the importance of handling noisy samples within dataset distillation. 
We believe this work will open up new avenues for practical and robust dataset distillation with noisy labels, making distilled datasets more applicable to a wide range of real-world applications.

\section*{Impact Statement}
This paper presents work whose goal is to advance the field of Machine Learning. 
There are many potential societal consequences of our work, but we do not feel that any of them must be specifically highlighted here.

\nocite{langley00}

\bibliography{main}
\bibliographystyle{conf}

\newpage
\appendix
\onecolumn

\end{document}